\documentclass[11pt]{article}

\usepackage[utf8]{inputenc}
\usepackage[T1]{fontenc}
\usepackage{amsmath,amssymb}
\usepackage{graphicx}
\usepackage{caption}
\usepackage{subcaption}
\usepackage{authblk}
\usepackage{geometry}
\usepackage{abstract}
\usepackage{hyperref}
\usepackage{tikz}

\title{\textbf{Hamiltonian Normalizing Flows as kinetic PDE solvers: application to the 1D Vlasov-Poisson Equations}}
\author[1]{Vincent Souveton}
\author[1]{Sébastien Terrana}
\affil[1]{CEA, DAM, DIF, Arpajon, F-91297, France\\

Corresponding Email: vincent.souveton@cea.fr}

\date{}  

\begin{document}

\maketitle

\begin{abstract}
\noindent Many conservative physical systems can be described using the Hamiltonian formalism. A notable example are the Vlasov-Poisson equations — a set of partial differential equations that govern the time evolution of a phase-space density function representing collisionless particles under a self-consistent potential. These equations play a central role in both plasma physics and cosmology. Due to the complexity of the potential involved, analytical solutions are rarely available, necessitating the use of numerical methods such as Particle-In-Cell. In this work, we introduce a novel approach based on Hamiltonian-informed Normalizing Flows, specifically a variant of Fixed-Kinetic Neural Hamiltonian Flows. Our method transforms an initial Gaussian distribution in phase space into the final distribution using a sequence of invertible, volume-preserving transformations derived from Hamiltonian dynamics. The model is trained on a dataset comprising initial and final states at a fixed time $T$, generated via numerical simulations. After training, the model enables fast sampling of the final distribution from any given initial state. Moreover, by automatically learning an interpretable physical potential, it can generalize to intermediate states not seen during training, offering insights into the system’s evolution across time.
\end{abstract}

\section{Introduction}

Complex particle systems appear across many areas of Physics, including plasma dynamics, astrophysics, and cosmology. In the mean-field limit — where the number of interacting particles becomes very large — these systems are often effectively described by kinetic equations derived as formal limits of their underlying microscopic behavior. These equations track the time evolution of the system's phase-space distribution. A prominent example is the Vlasov-Poisson system, which models the evolution of a distribution function under self-consistent fields and plays a central role in describing both collisionless plasmas and the formation of large-scale structures in the Universe \cite{nicholson1983introduction, bernardeau2002large}. Other examples include the Boltzmann and Landau equations, which provide effective descriptions of dilute or weakly-coupled particle ensembles \cite{cercignani1988boltzmann, villani2002review}. In this paper, we specifically aim at describing systems governed by Hamiltonian dynamics, which preserve the total energy of the system and encode fundamental symmetries of the underlying physics \cite{goldstein2002classical, arnold2013mathematical}. \\

Due to the high dimensionality and nonlinearity of these equations, analytical solutions are generally intractable. Consequently, numerical simulations become valuable tools for investigating the underlying dynamics. However, simulating such systems is computationally intensive. Direct $N$-body simulations, which compute pairwise interactions explicitly, scale with a complexity of $\mathcal{O}(N^2)$, making them impractical for large systems \cite{hockney1988computer}. To address this, Particle-In-Cell (PIC) methods offer a more efficient alternative by interpolating particle properties onto a grid to solve field equations, reducing the computational cost to $\mathcal{O}(N + N_g \ln N_g)$, where $N_g$ is the number of grid points \cite{birdsall1991plasma}. Monte Carlo methods, particularly stochastic particle approaches, are also used, especially when collisions or uncertainty quantification are involved \cite{nanbu1997theory}. A major drawback of these techniques is their cost: each new initial configuration typically requires rerunning a full simulation, which is computationally expensive. \\

Recently, generative models from Machine Learning have emerged as powerful surrogate tools for simulating high-dimensional dynamical systems. Once trained on a diverse set of data, probabilistic generative models such as Normalizing Flows \cite{reviewNF}, Variational Autoencoders \cite{VAE}, Generative Adversarial Networks \cite{GANs} and Diffusion models \cite{diffusionModels, DDPM} can rapidly generate statistically accurate samples from complex target distributions, achieving high fidelity and orders-of-magnitude speedup compared to traditional simulators. In the field of simulation, applications include  emulation of the cosmic web structure \cite{rodriguez2018fast} and surrogate modeling in high-energy physics \cite{paganini2018calogan}. Crucially, these methods are particularly well-suited when the system admits a probabilistic description such as a phase-space distribution. Thanks to their generalization properties, they allow one to bypass costly sampling via repeated simulation. \\

To further improve the physical consistency of generative models, recent advances have integrated domain knowledge directly into the learning process. Physics-Informed Neural Networks \cite{raissi2019physics} represent one such paradigm, where the neural network is trained not only on data but also by enforcing the residuals of the governing PDE directly in the loss function. More generally, operator-learning techniques like Fourier Neural Operators \cite{li2021fourier} aim to learn solution operators of Partial Differential Equations between function spaces, enabling generalization across initial conditions and geometries. In the context of Hamiltonian systems, where energy conservation and symplectic structure are central, deep learning architectures like Hamiltonian Neural Networks \cite{HNN} or Hamiltonian Generative Networks \cite{NHF} have been introduced to explicitly encode these invariants into the model architecture. They leverage the structure of classical Mechanics to generate physically plausible trajectories, and are particularly suited to modeling long-time dynamics of conservative systems. In the context of sampling, this led to the development of Hamiltonian-based Normalizing Flows \cite{NHF, EquivariantHF} that are able to perform high-dimensional sampling while being highly interpretable. \\

This work makes three primary contributions to the simulation of Hamiltonian kinetic systems:

\begin{itemize}
    \item We introduce \textbf{PDE-NHF}, a robust, Physics-informed Normalizing Flows model designed for fast and accurate integration of Hamiltonian kinetic equations in phase space over a fixed time interval $[0, T]$. The model is trained using a probabilistic objective function to capture the evolution of the full particle distribution. \\
    
    \item We construct an interpretable self-consistent potential function by employing an invariant function under specific symmetry groups, ensuring that the learned potential respects the underlying Physics of the system. \\
    
    \item Through a simple one-dimensional experiment, we demonstrate that PDE-NHF is capable of automatically recovering the correct Vlasov–Poisson dynamics throughout the entire trajectory $0 \leq t \leq T$. The model effectively learns an accurate approximation of the true Hamiltonian and exhibits robustness to the choice of the numerical integrator hyperparameters, enabling acceleration strategies for efficient simulation.
\end{itemize}

The paper is organized as follows. In Section~\ref{sec:RelatedWork}, we review related work. Section~\ref{sec:NHF} provides background on Normalizing Flows, with a focus on Hamiltonian-based models. In Section~\ref{sec:FKNHFforPDE}, we present \textbf{PDE-NHF}, our automatic integrator for the Vlasov–Poisson system. Section~\ref{sec:NumericalExperiments} illustrates the behavior of PDE-NHF on a one-dimensional example. Finally, Section~\ref{sec:Conclusion} concludes the paper with a discussion of the results and future directions.


\section{Related work}\label{sec:RelatedWork}

\noindent\textbf{Particle-In-Cell methods.} Solving kinetic equations is often achieved through numerically efficient PIC methods. For instance, WarpX \cite{WarpX}, OSIRIS \cite{OSIRIS} or Smilei \cite{Smilei} are PIC-based codes used to simulate plasma dynamics. In this paper, we design a straightforward PIC code for generating the training dataset. It is noteworthy that PIC methods can be combined with ML techniques for efficient computation of the electric field on the discretized grid \cite{ML-PIC}. Our approach is distinct in that it automatically learns the system’s Hamiltonian, enabling the straightforward incorporation of physical prior knowledge without relying on a binning strategy. Also, it is solely trained on initial and final states and it comes with the ability to extrapolate to intermediate timesteps. Finally, the training objective is formulated in terms of Kullback-Leibler divergence, focusing on the probabilistic descriptions of the system instead of the trajectory of each single particle. \\

\noindent\textbf{Generative models for kinetic equations.} Since modeling a plasma consists in simulating the evolution of a distribution in phase space, Normalizing Flows are naturally suited for this task. In \cite{CNFVlasov}, the authors use computationally-intensive Continuous Normalizing Flows \cite{NeuralODE, Grathwohl2019} for tracking the phase-space distribution along characteristic curves. In \cite{NFVlasov}, they employ pseudoreversible Normalizing Flows \cite{PseudoReversibleNN, SelfNF} as surrogate models for simulating the Fokker-Planck dynamics. Another approach is to rely on Physics-informed models \cite{PINNVlasovPoisson}. In our paper, we rely on Hamiltonian-informed Normalizing Flows for fast and accurate sampling of the final distribution in phase space without explicit knowledge of the Vlasov-Poisson equation, as well as learning the system's evolution between the initial and the final states. \\

\noindent\textbf{Learning Hamiltonians.} Our problem deals with learning the Hamiltonian of a specific physical system. In the literature, Hamiltonian Neural Networks \cite{HNN} and Hamiltonian Generative Networks \cite{NHF} have been successfully applied on the task of learning a Hamiltonian from data. In this work, we are rather interested in learning interpretable Hamiltonian transformations of probability distribution functions instead of following each point of the system individually. In this sense, our work is related to Hamiltonian flow-based models for sampling complex distributions known as Neural Hamiltonian Flows \cite{NHF, FKNHF}. The main differences lie in the fact that we are directly working in the full phase space and that the potential that needs to be learned is not fixed anymore as it depends on the current state of the system.


\section{Hamiltonian-based Normalizing Flows}\label{sec:NHF}

\subsection{Normalizing Flows}

Normalizing Flows \cite{densityEstimationLL, NICE, VIwithNF, reviewNF} are a powerful class of generative models that transform a simple base distribution $\pi_0 \in \mathcal{C}^\infty(\mathbb{R}^d)$ (typically a standard Gaussian) into a more complex target distribution $\pi \in \mathcal{C}^\infty(\mathbb{R}^d)$ through a finite sequence of smooth, invertible transformations. Formally, the overall transformation is denoted by:
$$ \mathcal{T}_\theta = \mathcal{T}_\theta^L \circ \cdots \circ \mathcal{T}_\theta^1,$$
where each $\mathcal{T}_\theta^\ell$, for $1 \leq l \leq L $, represents a learned bijective map, and $\theta$ denotes the collective set of parameters. Given a latent variable $Z \sim \pi_0$, the output variable is obtained as $Q = \mathcal{T}_\theta(Z)$. Using the change of variables formula, the model density $\pi_\theta$ is expressed as:
\[
\pi_\theta(q) = \pi_0\left( \mathcal{T}_\theta^{-1}(q) \right) \times \prod_{\ell=1}^L \left| \det \, \text{Jac}_{(\mathcal{T}_\theta^\ell)^{-1}}(q) \right|,
\]
where $\text{Jac}_{(\mathcal{T}_\theta^\ell)^{-1}}$ denotes the Jacobian matrix of the inverse transformation at step $\ell$. \\

The model is typically trained by minimizing the forward Kullback–Leibler divergence between the target distribution $\pi$ and the model distribution $\pi_\theta$, leading to the optimization objective:
\begin{align*}
&\mathcal{L}(\theta) = \mathbb{E}_{q \sim \pi} \left[ \ln \pi(q) - \ln \pi_\theta(q) \right] \\
&= -\mathbb{E}_{q \sim \pi} \left[ \ln \pi_0\left( \mathcal{T}_\theta^{-1}(q) \right) + \sum_{\ell=1}^L \ln \left| \det \, \text{Jac}_{(\mathcal{T}_\theta^\ell)^{-1}}(q) \right| \right] + C,
\end{align*}
where $C$ is a constant independent of $\theta$. In practice, the expectation is approximated via Monte Carlo sampling over training data. \\

Designing effective flow-based models requires choosing transformations $\mathcal{T}_\theta^\ell$ that satisfy several key properties: \textbf{(i)} they are smooth and easily invertible; \textbf{(ii)} their Jacobian determinant is easy to compute; \textbf{(iii)} the resulting chain of transformations is expressive enough to approximate a wide class of target distributions; \textbf{(iv)} ideally, the chain of transformation is interpretable. \\

\subsection{Hamiltonian Mechanics}

Hamiltonian Mechanics is a reformulation of classical Mechanics that describes the evolution of a system in terms of its generalized coordinates: positions $q \in \mathbb{R}^d$ and momenta $p \in \mathbb{R}^d$. The state of the system $(q(t), p(t))$ evolves according to a set of first-order differential equations, known as Hamilton's equations, which are applied componentwise:
\[
\left\{
\begin{aligned}
    \frac{dq}{dt} &= \frac{\partial H}{\partial p}, \\
    \frac{dp}{dt} &= -\frac{\partial H}{\partial q}.
\end{aligned}
\right.
\]
Here, $H = H(q, p)$ is the Hamiltonian function, typically representing the total energy of the system. In many cases, it can be decomposed into the sum of a potential energy term $V(q)$, depending only on the positions and a kinetic energy term $K(p)$, depending only on the momenta:
\[
H(q, p) = V(q) + K(p).
\]

Starting from an initial condition $(q_0, p_0)$ in phase space, Hamilton's equations define a unique flow $\mathcal{T}^t$ that evolves the state to $(q(t), p(t))$ at any future time $t > 0$. This flow possesses several fundamental properties:

\begin{itemize}
    \item \textbf{Time reversibility:} The dynamics are invariant under time reversal, meaning that running the system backward in time exactly retraces its trajectory. \\
    
    \item \textbf{Energy conservation:} The Hamiltonian remains constant along trajectories if it does not explicitly depend on time. \\
    
    \item \textbf{Symplecticity:} The flow $\mathcal{T}^t$ is symplectic, meaning that it preserves the symplectic form $\omega = dq \wedge dp$. In practical terms, this implies that the volume in phase space is conserved during the evolution, a property formalized by Liouville's theorem:
    \[
    \frac{d}{dt} \, \text{det}\left( \frac{\partial (q(t), p(t))}{\partial (q_0, p_0)} \right) = 0.
    \]
    Therefore, the Jacobian determinant of the flow remains exactly equal to one for all time.
\end{itemize}

These properties make Hamiltonian Mechanics particularly suitable for describing conservative physical systems, and they motivate the use of symplectic numerical integrators, such as the Leapfrog scheme, when solving Hamiltonian systems computationally. \\ 

Hamiltonian mappings combine the four above-mentioned properties because \textbf{(i)} they are smooth and easily invertible, as one just needs to reverse the timestep to go backward in time; \textbf{(ii)} as symplectic transformations, they are volume-preserving, meaning that their Jacobian determinant is exactly equal to 1; \textbf{(iii)} they have empirically demonstrated their ability to model complex distributions \cite{HMC, NHF, FKNHF}; \textbf{(iv)} they are Physics-inspired, making them highly interpretable in many cases \cite{FKNHF}.

\subsection{Fixed-Kinetic Neural Hamiltonian Flows}

Neural Hamiltonian Flows (NHF) \cite{NHF} are Physics-informed models that treat Hamiltonian transformations as normalizing flows. They are used to perform sampling from a target probability distribution with probability density function $f(q) \in \mathcal{C}^\infty(\mathbb{R}^d)$. In training mode, a point $q$ is sampled from the training dataset. It is seen as a position vector associated to a single particle in dimension $d$. To simulate Hamiltonian dynamics, one needs to add artificial momenta $p$. This is achieved by an Encoder made of two neural networks $\mu_\theta$ and $\sigma_\theta$ which are used to sample $p \sim p_\theta(q) = \mathcal{N}(\mu_\theta(q), \sigma_\theta(q)^2)$.\\

Then, one defines a kinetic energy $K_\theta$ as well as a potential energy $V_\theta$ for moving through phase space. They are usually modeled as neural networks, too. However, in \cite{FKNHF}, the authors propose to fix the kinetic term to a quadratic form as in classical Mechanics in order to improve interpretability and reduce the model complexity. In the resulting Fixed-Kinetic Neural Hamiltonian Flows model (FKNHF), one should thus set $K_\theta(p)=\frac{1}{2}p\mathcal{M}^{-1}_\theta p^T$ with $\mathcal{M}^{-1}_\theta$ a positive-definite mass matrix learned during training. As for the potential energy $V_\theta$, it is modeled as a neural network that depends on $q$. \\

System evolves in phase space following Hamilton's equations approximated by a symplectic Leapfrog integrator for $L$ steps with timestep $\delta t$ and initial conditions $(q(0), p(0))$. One step of the Leapfrog scheme reads:
\[
\left\{
\begin{aligned}
    p\left(t + \frac{\delta t}{2}\right) &= p(t) - \frac{\delta t}{2} \nabla_q V_\theta(q(t)), \\
    q(t + \delta t) &= q(t) + \ \delta t \ p\left(t + \frac{\delta t}{2} \right)\mathcal{M}^{-1}_\theta \\
    p(t + \delta t) &= p\left(t + \frac{\delta t}{2}\right) - \frac{\delta t}{2} \nabla_q V_\theta\left(q(t + \delta t) \right).
\end{aligned}
\right.
\]

The outputs $q(L\delta t)$ and $p(L\delta t)$ obtained after $L$ Leapfrog steps iterations are supposed to follow the joint base distribution $\Pi_0$. The model is trained by \textbf{maximizing} the following evidence lower bound \cite{NHF}:
$$\mathcal{L}_{\text{NHF}}(q;\theta) = \mathbb{E}_{p \sim p_\theta(q)} \left[ \ln \Pi_0\left( \mathcal{T}_\theta^{-1}(q, p) \right) - \ln p_\theta(p | q) \right].$$ 

Once the model is trained, the generative process involves drawing a pair $(q(0),p(0)) \sim \Pi_0$ that follows the joint base distribution, applying $L$ Leapfrog step with timestep $-\delta t$ and marginalizing over the momenta in the final state $(q(-L\delta t), p(-L\delta t))$ to get a sample $q(-L\delta t)$ that follows the learned target distribution.


\section{Solving Hamiltonian kinetic equations with PDE-NHF}\label{sec:FKNHFforPDE}

\subsection{Vlasov-Poisson system}

Hamiltonian kinetic partial differential equations describe the temporal evolution of a particle distribution $\Pi(q, p, t)$ in phase space, under the action of a Hamiltonian flow. One of the most fundamental examples is the \textbf{Vlasov equation}, which, in one dimension, reads:
$$\frac{\partial \Pi}{\partial t} + p \frac{\partial \Pi}{\partial q} + F(q,t) \frac{\partial \Pi}{\partial p} = 0.$$
This equation governs the evolution of a collisionless system subjected to a force field $F(q,t)$. The terms on the left-hand side correspond respectively to temporal evolution, spatial advection, and acceleration in momentum space. The structure of the Vlasov equation follows directly from Liouville's theorem \cite{LandauLifshitzMechanics}, which asserts that the probability density function $\Pi$ is conserved along the phase-space trajectories defined by Hamilton's equations. \\

In many applications, the force $F$ is not fixed externally but instead depends on the evolving distribution of particles, leading to a self-consistent coupling between the Vlasov equation and a field equation. Typically, the force derives from a potential $V(q,t)$ as:
$$F(q,t) = -\frac{\partial V(q,t)}{\partial q}.$$
The potential itself satisfies a second-order \textbf{Poisson equation}, whose form depends on the physical context:
\begin{itemize}
    \item \textbf{Cosmology} (self-gravitating systems):
    \[
    \frac{\partial^2 V}{\partial q^2} = 4\pi G \rho(q,t),
    \]
    where $G$ is the gravitational constant and $\rho(q,t) = \int \Pi(q,p,t) \, dp$ is the mass density of cold dark matter particles. \\
    
    \item \textbf{Plasma Physics} (electrostatic systems):
    \[
    \frac{\partial^2 V}{\partial q^2} = -\frac{\rho(q,t)}{\varepsilon_0},
    \]
    where $\varepsilon_0$ is the permittivity of the medium, and $\rho(q,t)$ is the charge density associated with the distribution of charged particles.
\end{itemize}

Thus, the Vlasov–Poisson system models the self-consistent evolution of particle ensembles under their collective gravitational or electrostatic interactions, while preserving the symplectic structure of Hamiltonian dynamics.

\subsection{Introducing PDE-NHF}

Originally, Hamiltonian-based Normalizing Flows were developed to perform efficient sampling in high-dimensional spaces. Since the goal is to learn an interpretable Hamiltonian map between two distributions in phase space, it is natural to reinterpret these models as solvers for Hamiltonian kinetic PDEs. In the following, $\mathbf{q} := (q_1,...,q_N) \in \mathbb{R}^{N \times d}$ and $\mathbf{p} := (p_1,...,p_N) \in \mathbb{R}^{N \times d}$ denote the generalized positions and momenta, respectively, of a $N$-particle system living in a $d$-dimensional space. We now formulate the Machine Learning problem we aim to address. \\

\textbf{Context.} We are given a collection of initial Gaussian distributions $\Pi_0^\mathbf{x}$ in phase space, each parametrized by a mean and covariance matrix encoded through a vector $\mathbf{x} \in \mathcal{D}$. Additionally, for each initial distribution, we are provided with the corresponding final set of particle positions $\mathbf{q}_T$ and momenta $\mathbf{p}_T$ at a fixed time $T > 0$, obtained by evolving a system of $N$ particles according to a Hamiltonian dynamics (\textit{e.g.} the one-dimensional Vlasov–Poisson equations). We aim to train a model that maps any initial Gaussian distribution $\Pi_0^\mathbf{x}$, $\mathbf{x} \in \mathcal{D}$, to the corresponding final distribution $\Pi_T^\mathbf{x}$. Moreover, the model should be capable of reconstructing physically consistent intermediate distributions $\Pi_t^\mathbf{x}$ for all $0 < t < T$. Importantly, we aim to correctly model the evolution of probability distributions, rather than tracking individual particle trajectories. \\

\textbf{Solution.} We propose a new generative model named \textbf{PDE-NHF}, derived from the FKNHF framework, with the following key features:
\begin{itemize}
    \item The total energy of the $N$-particle system is written as the sum of a fixed quadratic kinetic term and a self-consistent potential energy term. Specifically, the kinetic energy is given by:
    $$K_\theta(\mathbf{p}) = \frac{a^2}{2} \sum_{j=1}^N ||p_j||^2,$$
    where $a$ is a scalar parameter learned during training. The self-consistent potential energy $V_\theta(\mathbf{q})$ is parametrized by a neural network. The system evolves in phase space using a Leapfrog integrator, thus ensuring that the symplectic structure is directly embedded into the model's architecture rather than merely enforced through the loss function. \\
    
    \item Unlike traditional FKNHF approaches, PDE-NHF works directly in full phase space, eliminating the need for an Encoder to generate artificial momenta. The model distribution remains exactly tractable at any time $0 \leq t \leq T$. Given that the Jacobian determinant of the transformation is exactly $1$, the probability density evolves as
    $$\Pi_{\theta,t}^\mathbf{x}(\mathbf{q}_t, \mathbf{p}_t) = \Pi_0^\mathbf{x}\left( (\mathcal{T}_\theta^t)^{-1}(\mathbf{q}_t, \mathbf{p}_t) \right).$$
    Consequently, the training objective is to minimize the Kullback–Leibler divergence between the target and model distributions, conditioned on the initial distribution parameters encoded in $\mathbf{x}$. Given a training dataset $\{(\mathbf{q}_T, \mathbf{p}_T, \mathbf{x})_m\}_{m=1}^M$, the loss function is approximated by
    \begin{align*}
        \mathcal{L}_\text{PDE-NHF}(\theta) &= -\mathbb{E}_{(\mathbf{q}_T, \mathbf{p}_T) \sim \Pi_T^\mathbf{x}} \left[ \ln \Pi_0^\mathbf{x}\left( (\mathcal{T}_\theta^T)^{-1}(\mathbf{q}_T, \mathbf{p}_T) \right) \right] \\
        &\approx -\frac{1}{M} \sum_{m=1}^M \ln \Pi_0^\mathbf{x}\left( (\mathcal{T}_\theta^T)^{-1}(\mathbf{q}_T, \mathbf{p}_T) \right).
    \end{align*}
    Notably, this Monte Carlo estimation only requires access to the initial phase-space distribution parameters (mean and covariance) and does not depend on the explicit initial particle realizations. \\
    
    \item Enforcing the model to learn interpretable dynamics can be achieved in multiple ways. A first idea would consist in adding extra terms to the loss function representing the model ability to follow the correct dynamics at a few intermediate timesteps. Instead, we chose to impose geometric constraints based on symmetry considerations \cite{GeometricDeepLearning}. In particular, the self-consistent potential energy function $V_\theta$ is designed to be invariant under permutations and translations of particles. We achieve this by defining $V_\theta$ as a Deep Set network \cite{DeepSets}, where for an input $\mathbf{q}$, the output is given by:
    $$V_\theta(\mathbf{q}) = \psi\left( \sum_{j=1}^N \phi\left( q_j - \frac{1}{N} \sum_{l=1}^N q_l \right) \right),$$
    with $\psi$ and $\phi$ implemented as multilayer perceptrons (MLPs). This architecture guarantees the required invariance properties by construction.
\end{itemize}

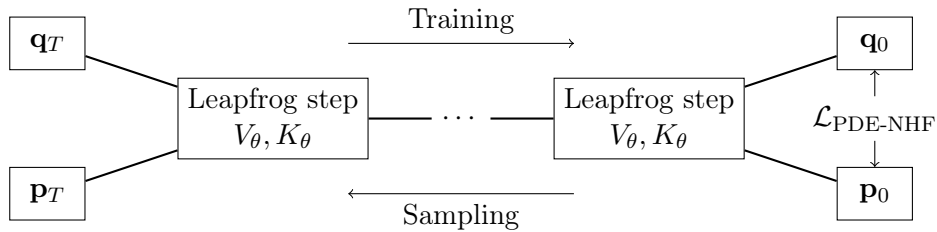
\begin{figure}[ht!]
\centering
\begin{tikzpicture}
  \node[draw, rectangle, minimum width=1.0cm, minimum height=0.7cm] (yt) at (-4, 1) {$\mathbf{q}_T$};
  \node[draw, rectangle, minimum width=1.0cm, minimum height=0.7cm] (pt_left) at (-4, -1) {$\mathbf{p}_T$};

  \node[draw, rectangle, minimum width=1.2cm, minimum height=1.0cm, align=center] (lf1) at (-1, 0) {Leapfrog step\\$V_\theta, K_\theta$};
  \node[draw, rectangle, minimum width=1.2cm, minimum height=1.0cm, align=center] (lf2) at (4, 0) {Leapfrog step\\$V_\theta, K_\theta$};
  \node (dots) at (1.5, 0) {$\cdots$};

  \draw[-, thick] (yt) -- (lf1);
  \draw[-, thick] (pt_left) -- (lf1);

  \node[draw, rectangle, minimum width=1.0cm, minimum height=0.7cm] (y0) at (7, 1) {$\mathbf{q}_0$};
  \node[draw, rectangle, minimum width=1.0cm, minimum height=0.7cm] (p0) at (7, -1) {$\mathbf{p}_0$};

  \draw[-, thick] (lf2) -- (y0);
  \draw[-, thick] (lf2) -- (p0);

  \draw[-, thick] (lf1) -- (dots);
  \draw[-, thick] (dots) -- (lf2);
  
  \node (loss) at (7,0) {$\mathcal{L}_{\text{PDE-NHF}}$};
  \draw[->] (loss) -- (y0);
  \draw[->] (loss) -- (p0);
  
  \draw[->] (0, 1) -- (3, 1) node[midway, above] {Training};
  \draw[<-] (0, -1) -- (3, -1) node[midway, below] {Sampling};


\end{tikzpicture}
\caption{Schematic representation of PDE-NHF.}
\label{fig:NHF}
\end{figure}

An illustration of the PDE-NHF model can be found in Figure~\ref{fig:NHF}.

\section{Numerical application to the 1D Vlasov-Poisson equation}\label{sec:NumericalExperiments}

\subsection{Dataset}

The whole dataset is made of $32768$ examples. 20000 examples are used for training, $6384$ for validation and $6384$ for testing. Each example consists in tracking the phase-space evolution of a $256$-particle system in dimension $d=1$ at times $t=0.0, \ 0.2, \ 0.4, \ 0.6, \ 0.8, \ 1.0$. To achieve this, we implement a PIC method on a regular 1D grid made of $128$ cells with box length $128$. For generating sufficient diversity, the initial positions are drawn from a multivariate Normal distribution $\mathcal{N}(64, \sigma_q^2I)$ and the initial velocities from another multivariate Normal distribution $\mathcal{N}(0, \sigma_p^2I)$, with $\sigma_q$ and $\sigma_p$ random scalars between $0.5$ and $1.5$. This implies that the initial phase-space distribution is different for every example in the dataset. The simulation process follows the loop described below for $L=25$ steps. 
\begin{enumerate}
    \item First, the density $\rho$ on the grid is evaluated using the Cloud-In-Cell method \cite{CIC}, a first-order interpolation scheme used in PIC simulations to map particle quantities onto a fixed grid. On the regular 1D grid with spacing $\delta x$, and grid points located at positions $x_i = i \delta x$, let us say we have a particle at position $q_p$, carrying a charge $c_p$. Let $i$ be the index of the grid point to the left of $q_p$ such that $x_i \leq q_p < x_{i+1}$. We define the normalized distance from the particle to the left grid point as
    $$\delta = \frac{q_p - x_i}{\delta x}, \quad \text{with } 0 \leq \delta < 1. $$
    The CIC weights assigned to the two nearest grid points are:
    $$w_i = 1 - \delta = 1 - \frac{q_p - x_i}{\delta x}, \quad
    w_{i+1} = \delta = \frac{q_p - x_i}{\delta x}.$$
    The particle's quantity $c_p$ is distributed to the grid as follows:
    $$
    \rho_i = \rho_i + \frac{c_p}{\delta x} \cdot w_i, \quad
    \rho_{i+1} = \rho_{i+1} + \frac{c_p}{\delta x} \cdot w_{i+1}.
    $$
    Here, $\rho_i$ represents the charge at grid point $i$, and the division by $\delta x$ converts the assigned quantity into a density. All the particles have the same charge $c_p=0.1$.\\

    \item Then, the electrostatic Poisson equation is solved in Fourier space:
    $$\frac{\partial^2 V}{\partial q^2} = -\frac{\rho}{\varepsilon_0} \ \longrightarrow \ \hat{V}_k = \frac{\hat{\rho}_k}{\varepsilon_0k^2}.$$
    Also, 
    $$F = -\frac{\partial V}{\partial q} \ \longrightarrow \ \hat{F}_k = -ik\hat{V}_k.$$ 
    This yields:
    $$\hat{F}_k = \frac{-ik\hat{\rho}_k}{\varepsilon_0 k^2}.$$
    In our simulations, we assume that $\varepsilon_0=1$. \\

    \item Finally, the system is evolved in phase space using one step of a symplectic Leapfrog scheme with integration timestep $\delta t = 0.04$ and periodic boundary conditions for the positions. Note that the initial conditions, the final time $T$ and the size of the box are chosen such that no particle has crossed the boundaries in our examples.
\end{enumerate}

\subsection{Numerical performance}

We now assess the numerical performance of PDE-NHF on the dataset described in the previous subsection. \\

For our experiments, the Deep Set architecture used to parametrize the potential energy function consists of two multilayer perceptrons (MLPs), $\phi$ and $\psi$, each containing a single hidden layer with 256 neurons and Softplus activations. The resulting model comprises approximately $130$k learnable parameters. Training is performed using $L=25$ Leapfrog steps with a fixed timestep $\delta t = 0.04$, over $200$ epochs with random minibatches of size $128$, and a fixed learning rate of $0.0003$ using the Adam optimizer \cite{Adam}. The full training procedure takes about $1$ hour on an NVIDIA A100 GPU~\copyright. \\

For comparison, we also report results obtained with a baseline model. This baseline consists of two MLPs, each with two hidden layers, totaling around $140$k parameters when combining the two networks. The former are trained to directly map the initial particle states in phase space — sorted in ascending order — onto the final position distribution and the final momentum distribution, respectively. In this setting, the models act as transformations from the initial cumulative distribution function (CDF) in phase space onto each marginal final CDF. Training the baseline model for $200$ epochs under the same optimizer, minibatch size, and learning rate takes approximately $2$ minutes on the same NVIDIA A100 GPU~\copyright. \\

To quantitatively evaluate the model performances, we use the one-dimensional Wasserstein distance. Given two samples $\mathbf{a} = (a_1, \ldots, a_N)$ and $\mathbf{b} = (b_1, \ldots, b_N)$, the Wasserstein distance $W_1(\mathbf{a}, \mathbf{b})$ is computed by first sorting the samples such that $a_{(1)} \leq \cdots \leq a_{(N)}$ and $b_{(1)} \leq \cdots \leq b_{(N)}$, and then evaluating:
$$W_1(\mathbf{a}, \mathbf{b}) = \sum_{j=1}^N |b_{(j)} - a_{(j)}|.$$

We summarize our findings and observations in the following conclusions.

\begin{enumerate}
    \item First, \textbf{our model serves as an efficient integrator} of the Vlasov–Poisson equation by accurately sampling the correct final distribution in phase space. This performance is quantitatively evaluated using the one-dimensional Wasserstein distance. As shown in Table~\ref{tab:performance}, PDE-NHF significantly outperforms the baseline MLP-based integrator. 
    \begin{table}[ht]
        \centering
        \caption{Averaged Wasserstein distance between the true and the generated samples at final time $t=1.0$ evaluated on the test dataset. The lower the better.}
        \begin{tabular}{c|c|c}
             & Mean $W_1$ for positions & Mean $W_1$ for momenta \\
             \hline
            PDE-NHF & $\mathbf{0.057}$ & $\mathbf{0.079}$ \\
            \hline
            Basic & $0.138$ & $0.138$
        \end{tabular}
        \label{tab:performance}
    \end{table}

    \item Second, \textbf{our model is capable of accurately interpolating} the dynamics throughout the entire trajectory. Notably, unlike traditional Physics-Informed Neural Networks, PDE-NHF does not require supervision at intermediate timesteps. After training, the learned parameter $a$, which parametrizes the kinetic term $K_\theta(\mathbf{p}) = \frac{a^2}{2} \sum_{j=1}^N p_j^2$, converges to a value of $0.99$, very close to the true unit value used to generate the training data. This result highlights the model’s ability to recover the correct Hamiltonian structure solely from pairs of initial and final states and to accurately integrate the Vlasov–Poisson dynamics over the full time interval $[0,T]$. By constraining the Physics of the system, one prevents the model to learn arbitrary dynamics between the initial and the final conditions. The extrapolation performance is summarized in Table~\ref{tab:extrapolation}. One may observe that the error slightly increases over time, due to a combination of numerical errors from the integrator and approximation errors from the learned Hamiltonian, resulting in progressive error amplification. 
    \begin{table}[ht]
        \centering
        \caption{Averaged Wasserstein distance between the true and the generated samples at intermediate times with PDE-NHF evaluated on the test dataset.}
        \begin{tabular}{c|c|c}
             & Mean $W_1$ for positions & Mean $W_1$ for momenta \\
             \hline
             $t=0.2$ & $0.007$ & $0.046$ \\
             \hline
             $t=0.4$ & $0.009$ & $0.033$ \\
             \hline
             $t=0.6$ & $0.021$ & $0.053$ \\
             \hline
             $t=0.8$ & $0.038$ & $0.062$
        \end{tabular}
        \label{tab:extrapolation}
    \end{table}

    \item Finally, \textbf{our model enables straightforward acceleration} of the integration process. Once trained, PDE-NHF is robust with respect to both the number of Leapfrog steps $L$ and the integration timestep $\delta t$, as long as the condition $L \times \delta t = T$ is satisfied. This allows the number of Leapfrog steps to be reduced to accelerate sampling of the full phase-space trajectory without sacrificing accuracy. As shown in Table~\ref{tab:acceleration}, using only $L=5$ Leapfrog steps yields results comparable to those obtained with $L=25$ steps. The main computational cost lies in the gradient evaluations required by the Leapfrog scheme, with a total of $L+1$ gradient computations per trajectory. Consequently, for sufficiently large $L$, reducing the number of steps by a factor of $n$ leads to an approximate $n$-fold acceleration of the sampling procedure. In our experiments, generating a sample at $t=1.0$ from an initial phase-space distribution using the PIC method takes approximately $0.02$ seconds. By contrast, using PDE-NHF with $L=5$ steps achieves a twofold reduction in wall-clock time, and this advantage is expected to become even more significant in higher-dimensional settings.
    \begin{table}[ht]
        \centering
        \caption{Averaged Wasserstein distance between the true and the generated samples at intermediate and final times with PDE-NHF evaluated on the test dataset. We have used $L=5$ and $\delta t=0.2$.}
        \begin{tabular}{c|c|c}
             & Mean $W_1$ for positions & Mean $W_1$ for momenta \\
             \hline
             $t=0.2$ & $0.008$ & $0.047$ \\
             \hline
             $t=0.4$ & $0.011$ & $0.035$ \\
             \hline
             $t=0.6$ & $0.021$ & $0.050$ \\
             \hline
             $t=0.8$ & $0.035$ & $0.069$ \\
             \hline
             $t=1.0$ & $0.054$ & $0.088$
        \end{tabular}
        \label{tab:acceleration}
    \end{table}    
\end{enumerate}

\begin{figure}[ht]
    \includegraphics[width=1.0\linewidth]{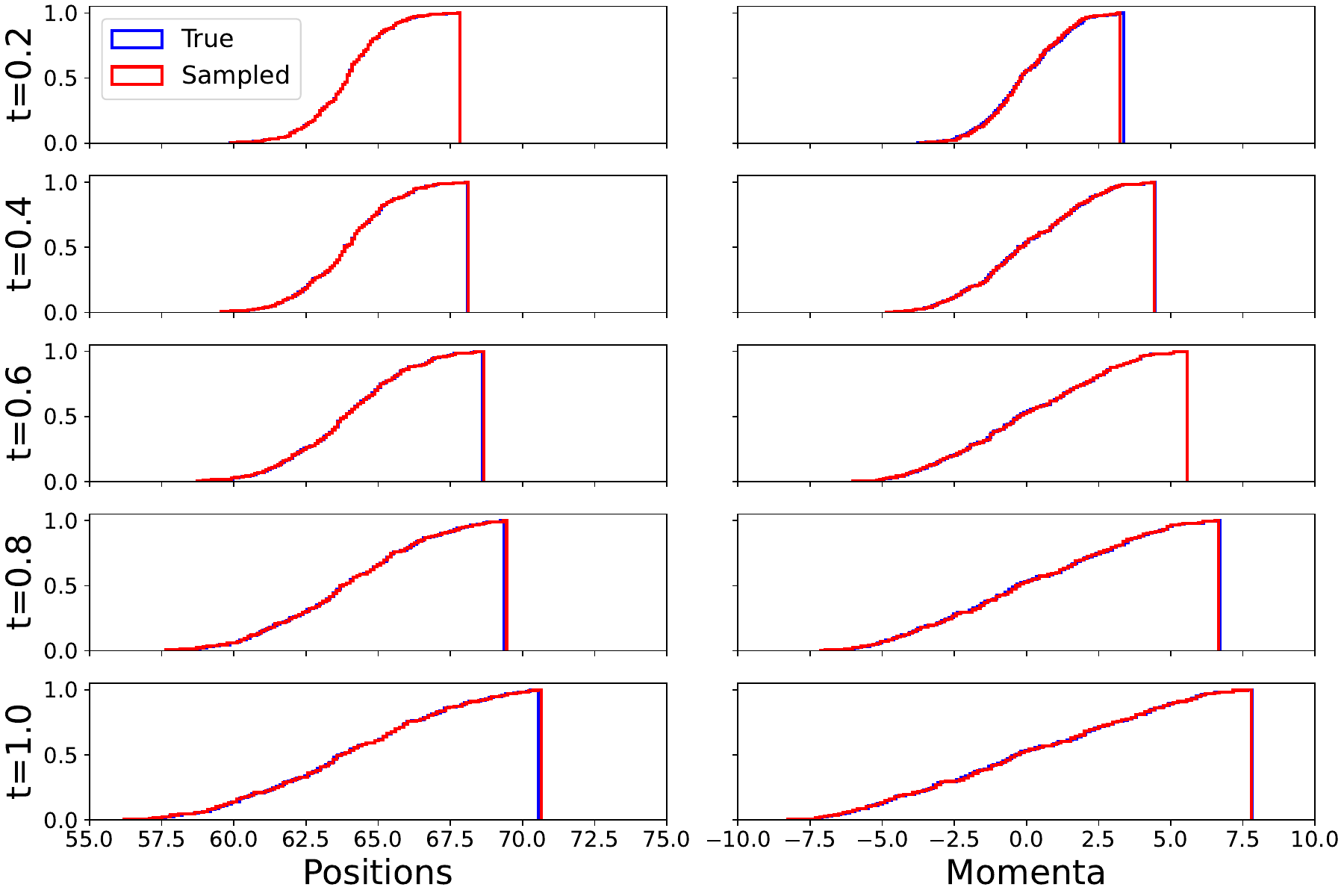}
    \caption{Normalized cumulative histograms from the true and the sampled trajectory in phase-space for one example from the test dataset. We used $L=25$ and $\delta t=0.04$ for both training and sampling.}
    \label{fig:example}
\end{figure}

We illustrate the capacity of PDE-NHF to integrate the correct dynamics along the whole interval $[0,T]$ on a test example in Figure~\ref{fig:example}. As can be seen from the cumulative histograms, our model fits the true phase-space distribution at final and intermediate timesteps with accuracy. As a consequence, the trained model may be seen as a fast and accurate surrogate for simulating the 1D Vlasov-Poisson dynamics. 


\section{Conclusion}\label{sec:Conclusion}

In this work, we introduced PDE-NHF, a Hamiltonian-based Normalizing Flows model designed to simulate particle systems governed by Hamiltonian kinetic equations. Leveraging a robust Leapfrog integration scheme, parametrized by a translation and permutation-invariant potential energy network, our generative model enables fast and accurate sampling of entire trajectories in phase space. Through experiments on a 1D Vlasov-Poisson system, we demonstrated the model’s ability to faithfully reconstruct the true Hamiltonian dynamics from pairs of initial and final phase-space states. Promising directions include extending our methodology to higher-dimensional, real-world datasets. For example, by training on samples representing the early phase-space distribution of the universe and their evolved states obtained through the integration of a Vlasov-Poisson system, our approach could provide a fast, accurate, and compact surrogate model for gravitational processes, capturing the evolution from homogeneous initial conditions to today’s complex filamentary cosmic structure.



\section*{Acknowledgments}
All authors are grateful to the Centre de Calcul Recherche et Technologie (CCRT) for providing computing and storage resources. 


\bibliographystyle{plain}  

\end{document}